\documentclass[letterpaper, 10 pt, conference]{ieeeconf} 
\IEEEoverridecommandlockouts
\usepackage{caption}
\usepackage{cite}
\usepackage{amsmath,amssymb,amsfonts}
\usepackage{algorithmic}
\usepackage{graphicx}
\usepackage{textcomp}
\usepackage{xcolor}
\usepackage{verbatim}
\let\labelindent\relax
\usepackage{enumitem}
\usepackage{algorithm}
\usepackage{array}
\usepackage{breqn}
\usepackage{tablefootnote}
\usepackage{hyperref}
\usepackage{subcaption}

\makeatletter
\newenvironment{breakablealgorithm}
  {
   \begin{center}
     \refstepcounter{algorithm}
     \hrule height.8pt depth0pt \kern2pt
     \renewcommand{\caption}[2][\relax]{
       {\raggedright\textbf{\fname@algorithm~\thealgorithm} ##2\par}%
       \ifx\relax##1\relax 
         \addcontentsline{loa}{algorithm}{\protect\numberline{\thealgorithm}##2}%
       \else 
         \addcontentsline{loa}{algorithm}{\protect\numberline{\thealgorithm}##1}%
       \fi
       \kern2pt\hrule\kern2pt
     }
  }{
     \kern2pt\hrule\relax
   \end{center}
  }
\makeatother

\newcolumntype{P}[1]{>{\centering\arraybackslash}p{#1}}
\newtheorem{remark}{Remark}
\newtheorem{proposition}{\textbf{Proposition}}

\newcommand{\calC}{\ensuremath{\mathcal{C}}}

\newcommand{\calI}{\ensuremath{\mathcal{I}}}
\newcommand{\calJ}{\ensuremath{\mathcal{J}}}

\newcommand{\calM}{\ensuremath{\mathcal{M}}}

\newcommand{\calP}{\ensuremath{\mathcal{P}}}

\begin{document}

\title{Dual-Arm Whole-Body Motion Planning: Leveraging Overlapping Kinematic Chains}

\author{Richard Cheng, Peter Werner, and Carolyn Matl
 \thanks{$^{1}$Toyota Research Institute}%
}
\maketitle
\thispagestyle{empty}
\pagestyle{empty}

\begin{abstract}
High degree-of-freedom dual-arm robots are becoming increasingly common due to their morphology enabling them to operate effectively in human environments. However, motion planning in real-time within unknown, changing environments remains a challenge for such robots due to the high dimensionality of the configuration space and the complex collision-avoidance constraints that must be obeyed. In this work, we propose a novel way to alleviate the curse of dimensionality by leveraging the structure imposed by shared joints (e.g. torso joints) in a dual-arm robot. First, we build two dynamic roadmaps (DRM) for each kinematic chain (i.e. left arm + torso, right arm + torso) with specific structure induced by the shared joints. Then, we show that we can leverage this structure to efficiently search through the composition of the two roadmaps and largely sidestep the curse of dimensionality. Finally, we run several experiments in a real-world grocery store with this motion planner on a 19 DoF mobile manipulation robot executing a grocery fulfillment task, achieving $\approx$0.4s average planning times with 99.9\% success rate across more than 2000 motion plans.
\end{abstract}

\section{Introduction}

High degree-of-freedom (DoF) dual-arm robots are increasingly deployed in manufacturing, healthcare, and service applications, due to their enhanced dexterity and similarity to human morphology. However, motion planning for such systems presents significant challenges due to the high-dimensional configuration space, complex collision-avoidance constraints, and the need to coordinate between arms. Due to the curse of dimensionality, these robots have extreme difficulty planning in changing environments in planning times adequate for real-world deployment. 

Current optimization-based approaches can deal with high DoF robots, but they struggle to operate robustly in changing environments due to the need to account for the complex non-convex collision-avoidance constraints arising from a large number of robot/environment collision geometries \cite{Yang2019}. Sampling-based approaches can effectively deal with changing environments by leveraging parallel computation \cite{Murray2016} and/or offline computation \cite{Kallmann2004}. However, they rely on simplification of the configuration space (e.g. projection to lower dimensions) to overcome the curse of dimensionality, which can lead to sub-optimality and planning failures, and/or require complex hand-crafted heuristics.

\begin{figure}[t]
\centering
\includegraphics[width=\columnwidth]{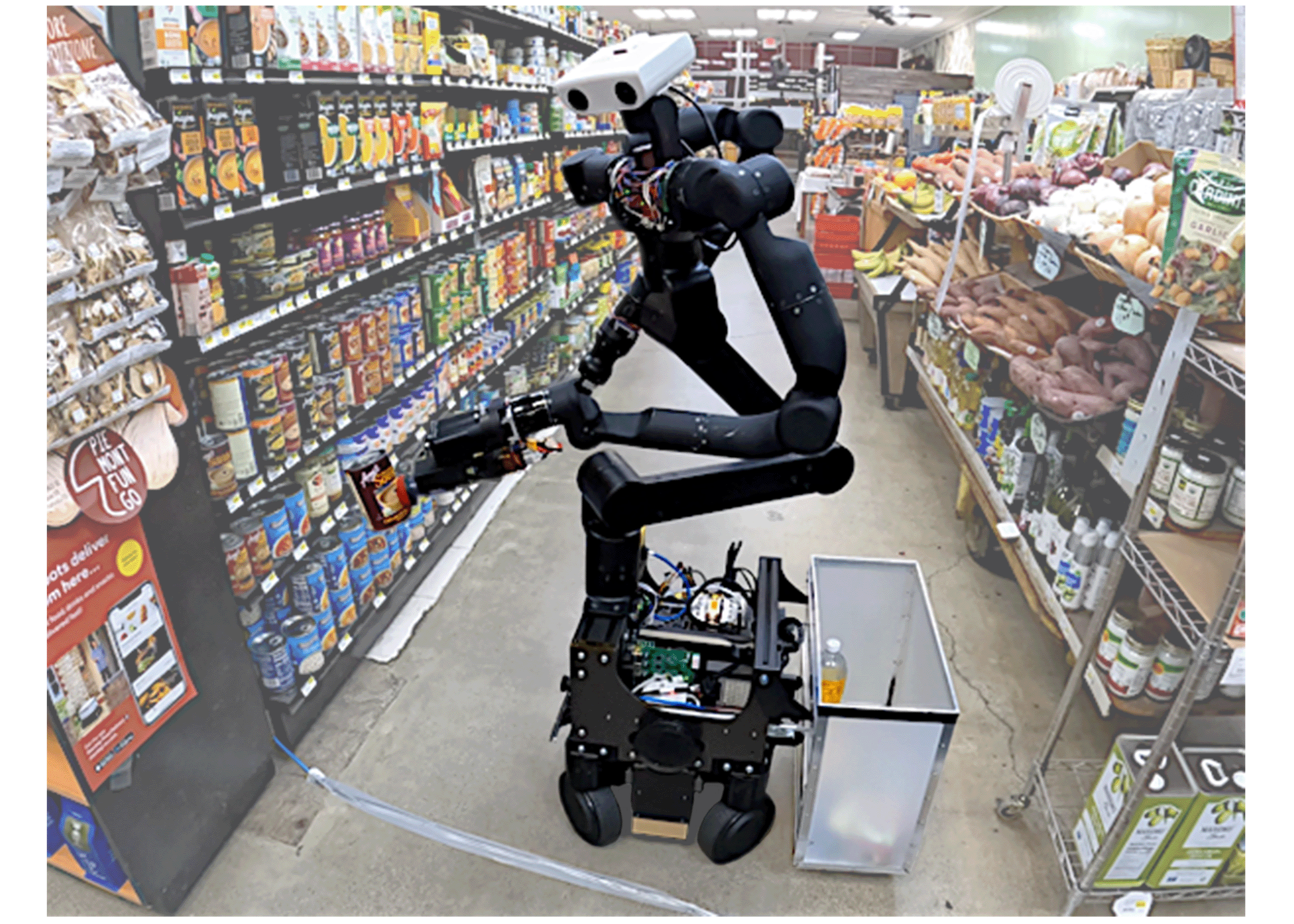}
\caption{Image of the robot grasping a can in the grocery store. The robot has a wheeled base, a 5 DoF torso, and two 7 DoF arms.}
\label{fig:robot_image}
\end{figure}

Our proposed approach alleviates the curse of dimensionality for high DoF dual-arm robots by leveraging the structure arising from their dual-arm nature. Dual-arm robots by design are composed of two kinematic chains with shared degrees of freedom at the ``root'' of the chains. For example, the robot pictured in Figure \ref{fig:robot_image} has a 5 DoF torso and two 7 DoF arms, resulting in 19 total DoFs; however, each kinematic chain for the left/right arms has 12 DoFs with the torso joints shared across both kinematic chains. This introduces significant structure into the motion planning problem. In this paper, we explore how we can intelligently leverage this structure to compose lower-dimensional dynamic roadmaps (DRM) from two kinematic chains in order to plan effectively for an overall higher-dimensional system. We find that our composition allows for greater efficiency even when the naive Cartesian product of the two graphs would make the problem \textit{worse}.

There are three main contributions of this paper:
\begin{itemize}
    \item A framework for building/composing multiple DRMs (for multiple kinematic chains) that leverages the natural decomposition of dual-arm systems with shared joints.
    \item A novel motion planning algorithm that can efficiently search for successful paths within our composition of roadmaps.
    \item Demonstration of our motion planning algorithm on a dual arm mobile manipulator with 19 DoFs performing a real task requiring complex motions in changing environments.
\end{itemize}

A video demonstrating the capabilities of this planning algorithm for our robot operating in a real grocery store can be found here: \url{https://drive.google.com/file/d/1CGAwj-Cti35Q1_XXdAs26Ba-LIGW_EF4/view?usp=drive_link}.

\section{Related Work}
\label{sec:related_work}

Most work in whole-body motion planning can be categorized into two classes: (1) sampling-based approaches \cite{Kuffner2000}, or (2) optimization-based approaches \cite{Zucker2013}, with many recent works leveraging a combination of the two \cite{Nguyen2016,Dai2018,Bajracharya2020, Werner2025}. Learning-based motion planners have also shown significant progress, but typically fall within one of the above two classes (e.g. learning a sampling heuristic \cite{Qureshi2019}). 

\subsection{Sampling-based motion planning}
\label{sec:sampling_based}

Sampling-based approaches rely on building a graph of robot joint configurations such that planning from one robot configuration to another reduces to a graph search problem. Rapidly exploring random trees (RRT) build this tree online when planning \cite{Kuffner2000}. The challenge then is intelligently choosing nodes to sample for the tree \cite{Gammell2014,Gammell2015,Panasoff2025} and efficiently expanding nodes to reduce expensive collision checks \cite{Hauser2015,Haghtalab2018}. Such methods have also had rapid speed-ups in recent years, leveraging intelligent hardware acceleration \cite{Thomason2024, Le2025}. 

In static environments, probabilistic roadmaps (PRM) can be built offline, such that online planning reduces to connecting the current configuration to the existing roadmap, searching through the roadmap for the node closest to the target configuration, and then connecting this node to the target configuration \cite{Kavraki1996}. Much work has dealt with optimal roadmap sampling strategies (e.g. to deal with narrow corridors or other challenging environments) \cite{Lavalle2006,Hsu2006,Chen2021}. However if the environment changes, then the PRM is no longer valid, as previously collision-free edges may now result in collision. Many methods have been proposed to alleviate this issue, for example using lazy collision checking and roadmap repair \cite{Jaillet2004} or storing similar obstacle maps \cite{Lien2010}.

To deal with dynamic environments, dynamic roadmaps (DRM) build both a traditional roadmap as well as a collision mapping from the voxelized workspace to nodes/edges of the roadmap \cite{Leven2002,Kallmann2004,Liu2006}. This mapping can then be used to update the roadmap when voxels are added/removed from the environment. However, due to the computational burden of generating/storing these DRMs, they have been limited to small roadmaps and low DoF systems \cite{Murray2016}. 



\subsection{Optimization-based motion planning}
\label{sec:optimization_based}

Optimization-based approaches formulate the motion planning problem as an optimization problem \cite{Kalakrishnan2011,Zucker2013}. Motion planning can be defined as a mixed-integer linear program (MILP), where the mixed integers arise due to the selection of the obstacle avoidance constraints (or other constraints) \cite{Ding2011}. However, mixed-integer programs are computationally intensive to solve and do not scale well. Alternatively, an initial seed path can be provided to the optimization, and any constraints are linearized around this path; therefore, the optimization problem can be approximately solved through sequential quadratic programming \cite{Schulman2014,Ichnowski2020gomp}. These solutions can scale better to higher degrees of freedom, but they suffer from brittleness and sensitivity to hyperparameters of the optimization or initial seed, which can lead to invalid and/or constraint-violating solutions \cite{Yang2019,Bajracharya2020}. A promising approach to overcome this has been to explore massive parallelization over seeds or hyperparameters to generate motion plans \cite{Sundaralingam2023}.

The Graphs of Convex Sets (GCS) planner \cite{Marcucci2022} formulates a convex relaxation of the mixed integer planning problem that is very tight and is able to achieve near-optimal motion plans. It allows bypassing non-convex collision-avoidance constraints, by generating collision-free convex sets that the robot must remain within. Significant progress has been made on speeding up this process \cite{Marcucci2024,werner2024faster, Werner2025}. However, the time to construct the convex sets is still a bottleneck, either limiting online planning in changing environments or requiring a valid path initialization. 

\subsection{Learning-based motion planning}

Learning-based motion planners have also shown significant progress and are typically tied to an underlying optimization-based or sampling-based planner in order to take advantage of their reliability and guarantees \cite{Ichter2018}. For example, many works have examined using supervised learning or reinforcement learning to learn an optimal sampling heuristic \cite{Zhang2018,Cheng2020,Arias2021,Qureshi2021}, or speed up collision-checking \cite{Das2020}. Other works have looked at planning in a lower-dimensional learned latent space \cite{Ichter2019, Yamada2023} or leveraging diffusion to learn trajectories \cite{Luo2024}.

\section{Problem Setup}

Consider the dual-arm redundant robot pictured in Figure \ref{fig:robot_image}. Given its current configuration $x_{start} \in \mathcal{C}$ and a target configuration $x_{target} \in \mathcal{C}$, our goal is to find a collision-free path that takes the robot from the current configuration to the final configuration. To aid in describing our problem, let us define the following:

\begin{itemize}[leftmargin=*]
  \item[] $\mathcal{C}_{arm_1}$ -- Configuration space for arm$_1$.
  \item[] $\mathcal{C}_{arm_2}$ -- Configuration space for arm$_2$.
  \item[] $\mathcal{C}_{torso}$ -- Configuration space for torso (shared DoFs).
  \item[] $\mathcal{C}$ -- Full configuration space ($\mathcal{C}_{arm_1} \times \mathcal{C}_{arm_2} \times \mathcal{C}_{torso}$).
  \item[] $\tau: [0,1] \rightarrow \mathcal{C}$ -- Parameterized robot path such that $\tau{(0)} = x_{start}$ and $\tau{(1)} = x_{target}$.
 \item[] $\mathbb{V}$ -- Voxel map space.
  \item[] $COL: \mathcal{C} \times \mathbb{V} \rightarrow \{0, 1\}$ -- Collision checker.
\end{itemize}

Given a voxel map $V \in \mathbb{V}$ and target configuration $x_{target}$, our goal is to solve the following problem, yielding a collision-free path $\tau$:

\begin{equation}
\begin{aligned}
\min_{\tau} \quad &  \int_{0}^{1} \| \tau(s) - x_{start}  \| ds \\
\textrm{s.t.} \quad & \tau{(0)} = x_{start} \\
  & \tau{(1)} = x_{target} \\
  & \neg COL(\tau(s), V) ~~~ \forall ~ s \in [0, 1]
\end{aligned}
\label{eq:optimization}
\end{equation}

Because of the collision-avoidance constraints, solving this optimization directly would require solving a high-dimensional mixed integer program -- an infeasible approach for online planning. Many approaches attempt to solve the problem approximately (e.g. sequential convex programming, semi-definite programming, etc.) but suffer from the pitfalls described in Section \ref{sec:optimization_based}. We will be exploring a sampling-based approach based on DRMs to solve this problem. 

\textbf{Voxelization: } In many practical mobile manipulation implementations, the collision environment is represented as a dynamic 3D voxel map, generated either through RGB-D cameras or LIDAR \cite{Bajracharya2023}. This voxel map allows us to represent dynamic environments as the robot moves, and environment collision is defined as intersection between any voxel and any robot collision body. Dealing with this dynamically changing voxel map can be dealt with by leveraging dynamic roadmaps (see Figure \ref{fig:drm}).

\section{Preliminaries - Dynamic Roadmaps}
\label{sec:drm}

\begin{figure}
    \centering
    \includegraphics[width=\linewidth]{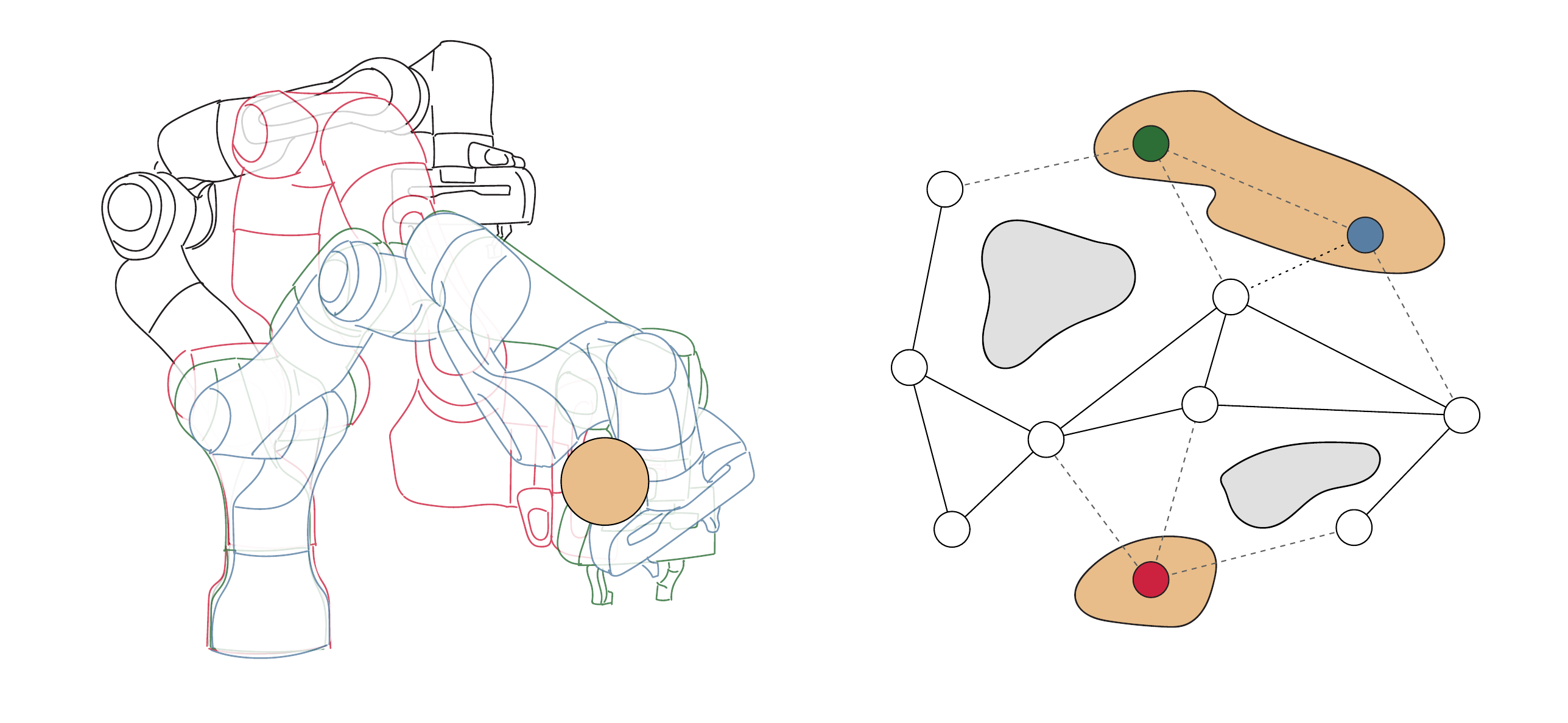}
    \caption{
    A DRM extends a PRM by adding lookup tables that enable rapid updates when obstacles appear in the task space. The task space is discretized into voxels, each represented by a circumscribing sphere for collision detection. A lookup table $\calM_c$ maps each sphere to roadmap nodes that would collide with it. The illustration shows a yellow sphere with three colliding configurations (red, green, blue) stored in $\calM_c$. The configuration space diagram displays the underlying PRM and obstacles, with the three colliding configurations highlighted and the yellow sphere's corresponding configuration obstacle marked in yellow. Figure adapted from \cite{Werner2025}.    
    }
    \label{fig:drm}
\end{figure}

Dynamic roadmaps were first proposed in 2002 in \cite{Leven2002} to enable planning in dynamic environments, but were initially considered impractical due to large memory/compute requirements. They have gained renewed interest in the past decade as memory/compute capabilities have exploded. DRMs approximate $\mathcal{C}_{free}$ with a PRM \cite{Kavraki1996}; in addition, they discretize the robot workspace into voxels and build large lookup tables that map ``active'' voxels to nodes/edges of the PRM that are in collision. The key idea is that pruning the PRM given a set of active voxels is faster and leads to better motion plans than running a single-query motion planner \cite{Kallmann2004} given the current perceived environment. Hence, DRMs enable us to offload much of the computational burden of motion planning (e.g. collision checks) to offline construction of the roadmap, and thus enable fast planning in \textit{dynamic} environments when voxels change. See Figure \ref{fig:drm} for a visual description of the DRM.

Our DRMs closely follow \cite{Werner2025} and consist of the following three data structures:
\begin{itemize}
    \item A node map $\calM_n: \calI\rightarrow\calC$ that maps a node identification number (node id) $i\in\calI$ to its configuration,
    \item a node adjacency map $\calM_a: \calI\rightarrow\calP(\calI)$, where $\calP$ denotes the power set of $\calI$, which maps a node id to all its neighbors, 
    \item a collision map $\calM_c: \calJ \rightarrow \calP(I)$ that maps a voxel identification number (voxel id) $j\in\calJ$ to all node ids that are in collision if the voxel $j$ is active.
\end{itemize}

We can then use these three data structures to rapidly find piecewise-linear paths online. Given a voxel map representation of the world, a starting configuration $x_{start}$, and a target configuration $x_{target}$, the planning proceeds in three phases. 
\begin{itemize}
    \item Build the collision set $CS\in\calP(I)$ that contains the ids of all nodes that are colliding with the voxel map. This is done by determining all voxels containing points and subsequently using the collision map $\calM_c$ to find all nodes in collision.
    \item Connect $x_{start}$ and $x_{target}$ to the closest nodes in the roadmap, $x^{\mathcal{R}}_{start}$ and $x^{\mathcal{R}}_{target}$ 
    \item Use $A^*$ combined with lazy collision checking and greedy short cutting to find the shortest collision-free path from $x^{\mathcal{R}}_{start}$ to $x^{\mathcal{R}}_{target}$.
\end{itemize}

The right side of Figure \ref{fig:drm} depicts the collision map being used to quickly prune out nodes in collision. The resulting collision-free roadmap can be efficiently used for path planning. The DRM described in this section will form the backbone of our approach, which relies on (1) imposing certain structure on the roadmap to allow composition of DRMs, and (2) enabling efficient search through this composition of roadmaps. Further details on the construction process for a single DRM can be found in \cite{Cheng2023}.

\section{Methods}

The ideal scenario would be to build a single DRM in the full configuration space $\mathcal{C}$ (19 DoFs). Then we could find a path from $x_{start}$ to $x_{target}$ within this DRM using $A^*$ and lazy collision-checking. Unfortunately, leveraging a DRM in this high-dimensional space is infeasible due to the curse of dimensionality (e.g. $5^{19}>1.9e13$ would require more than a petabyte of storage alone).

However, we can take advantage of the fact that the configuration space for any single kinematic chain in the system has significantly lower dimension, $dim(\mathcal{C}_{torso} \times \mathcal{C}_{arm})$, with $\mathcal{C}_{torso}$ being a shared configuration space. Our proposed algorithm is centered on leveraging this structure. Instead of trying to build/utilize a single massive DRM in the full configuration space, let us build and utilize two (connected) DRMs $R_{arm_1}, R_{arm_2}$ for the two kinematic chains, where $\mathcal{R}_{arm_1} \subset \mathcal{C}_{torso} \times \mathcal{C}_{arm_1}$ and $\mathcal{R}_{arm_2} \subset \mathcal{C}_{torso} \times \mathcal{C}_{arm_2}$. 

In contrast to previous decomposition approaches, we do not use a leader-follower approach for the two arms; this enables us to better leverage the shared joints for \textit{both} arms. Section \ref{sec:building_roadmaps} looks at how we build our two connected DRMs such that they can be efficiently composed in path planning and Section \ref{sec:planning_in_roadmaps} describes our online motion planning algorithm for efficient simultaneous search within the composed roadmaps.


\subsection{Building Composable Roadmaps (offline)}
\label{sec:building_roadmaps}

If we build $\mathcal{R}_{arm_1}$ and $\mathcal{R}_{arm_2}$ naively, we could end up planning in an implicit graph formed by the Cartesian product of our two graphs, making the dimension effectively $dim(\mathcal{C}_{torso} \times \mathcal{C}_{arm_1} \times \mathcal{C}_{torso} \times \mathcal{C}_{arm_2})$. This would make our problem \textit{much worse}. Therefore, we must ensure certain structure in our DRMs in order to efficiently compose them. 

Primarily, this simply entails adding the constraint that the vertices of both DRMs cover the \textit{exact} same subset of $\mathcal{C}_{torso}$. Let us denote the support of our DRM, $supp(\mathcal{R})$, as the set of all vertices in its roadmap, and let $supp_{\mathcal{C}}(\mathcal{R})$ denote the set of all of these vertices projected to subspace $\mathcal{C}$. Then we need only ensure that,
\begin{equation}
\textnormal{supp}_{\mathcal{C}_{torso}}(\mathcal{R}_{arm_1}) = \textnormal{supp}_{\mathcal{C}_{torso}}(\mathcal{R}_{arm_2}).
\end{equation}
The underlying concept is extremely simple and can be summarized as follows: if a certain torso configuration exists in $\mathcal{R}_{arm_1}$, then it must also exist in $\mathcal{R}_{arm_2}$, and vice versa. This can be enforced by construction if we consider a uniform discretization of the configuration spaces. By using the same uniform discretization of $\mathcal{C}_{torso}$ for both $\mathcal{R}_{arm_1}$ and $\mathcal{R}_{arm_2}$, we can ensure that their nodes span the same subset of the shared subspace $\mathcal{C}_{torso}$.

Unfortunately, there is no free lunch; clearly, we must sacrifice something in this reduction from 19D to 12D. In this case, $\textnormal{supp}(\mathcal{R}_{arm_1})$ has no knowledge of $\mathcal{C}_{arm_2}$, and vice versa. In other words, the arms do not know about each other! If a pair of vertices from $\mathcal{R}_1$ and $\mathcal{R}_2$ are in collision with a voxel or the torso, this would be encoded in the roadmap. However, if the arms are in collision with each other, this is not encoded. Therefore, during the roadmap construction process, we build another data structure:
\begin{itemize}
    \item An inter-arm collision map $\mathcal{M}_{inter_1}: \mathcal{I}_{arm_1} \rightarrow \mathcal{P}(\mathcal{I}_{arm_2})$, which maps nodes in $\mathcal{R}_{arm_1}$ to all node ids in $\mathcal{R}_{arm_2}$ that would result in the arms in collision.
    \item In the other direction: $\mathcal{M}_{inter_2}: \mathcal{I}_{arm_2} \rightarrow \mathcal{P}(\mathcal{I}_{arm_1})$
\end{itemize}

\begin{algorithm}[tb]
	\caption{Dual DRM Construction}\label{alg:roadmap_construction}
	\begin{algorithmic}[1]
        \STATE Uniformly discretize $\mathcal{C}_{torso}$, creating node set $\mathbb{V}_{Ctorso}$.
        \STATE Create $\mathbb{V}_{Carm_1}$ and $\mathbb{V}_{Carm_2}$ (optionally using uniform discretization).
        \STATE Generate the initial roadmaps from the created node sests: $\textnormal{supp}(\mathcal{R}_{arm_i}) = \mathbb{V}_{Ctorso} \times \mathbb{V}_{Carm_i}$.
	    \STATE Discard vertices that violate user-specified constraints (e.g. self-collision, COM bounds, etc.)
        \STATE From the remaining vertices, build the two DRMs $\mathcal{R}_{arm_1}$ and $\mathcal{R}_{arm_2}$ per the process outlined in Section \ref{sec:drm}, creating $\mathcal{M}_n, \mathcal{M}_a$, and $\mathcal{M}_c$ for each.
		\STATE Build the inter-arm collision map $\mathcal{M}_{inter}$, as described in Section \ref{sec:building_roadmaps}.
	\end{algorithmic}
\end{algorithm}

\begin{proposition}
    A node pair $(x_{\mathcal{R}_1}, x_{\mathcal{R}_2})$ (where $x_{\mathcal{R}_i} \subset \textnormal{supp}(\mathcal{R}_{arm_i})$) is in collision if and only if at least one of the following holds:
    \begin{itemize}
        \item $(x_{\mathcal{R}_1}, x_{\mathcal{R}_2})$ is contained in the inter-arm collision maps $\mathcal{M}_{inter_1}$ or $\mathcal{M}_{inter_2}$,
        \item $x_{\mathcal{R}_1}$ is in collision,
        \item $x_{\mathcal{R}_2}$ is in collision.
    \end{itemize}
\end{proposition}

This proposition is critical as it ensures that our DRM pair, $\mathcal{R}_{arm_1}, \mathcal{R}_{arm_2}$ contains all collision information that would be contained in the original 19D configuration space! See Appendix for proof. With this, we can be confident our two DRMs now encode all the information needed (accumulated offline) to enable valid, efficient online planning. Algorithm \ref{alg:roadmap_construction} outlines the DRM construction process.

\subsection{Planning within Composition of Roadmaps (online)}
\label{sec:planning_in_roadmaps}

\begin{figure*}
    \hfill
    \includegraphics[width=0.84\linewidth]{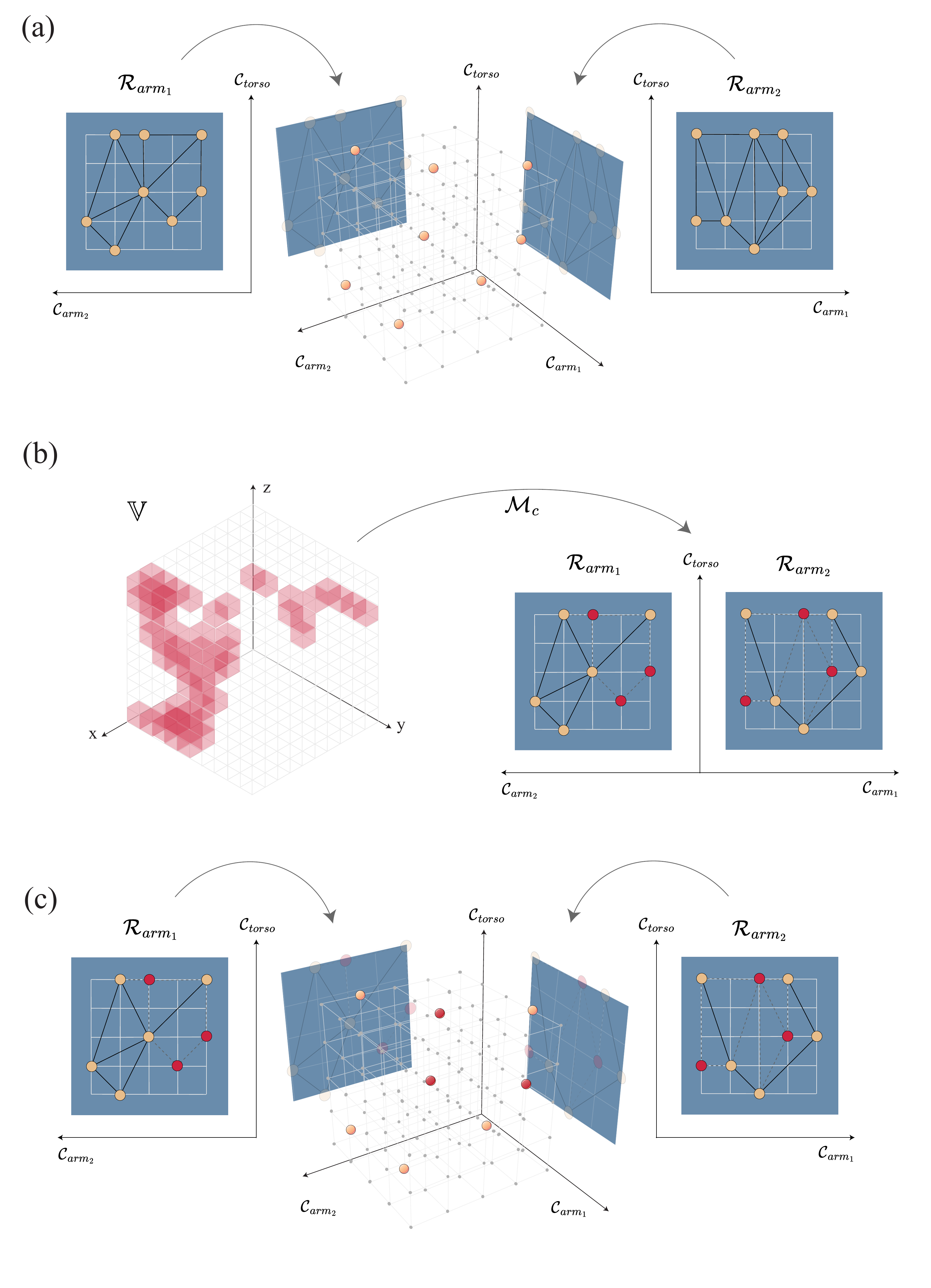}
    \caption{
(a) This depicts a 3D representation of the high-dimensional space spanned by $\mathcal{C}_{torso} \times \mathcal{C}_{arm_1} \times \mathcal{C}_{arm_2}$. The 2D projection on the left represents $\mathcal{R}_{arm_1}$, the DRM for kinematic chain 1, and the 2D projection on the right represents, $\mathcal{R}_{arm_2}$, the DRM for kinematic chain 2. In the center, the 2D projections for each kinematic chain are composed to form an \textit{implicit} 3D graph spanning $\mathcal{C}_{torso} \times \mathcal{C}_{arm_1} \times \mathcal{C}_{arm_2}$ with $\mathcal{C}_{torso}$ shared by the two chains. (b) Once a voxel map is provided, nodes in $\mathcal{R}_{arm_1}$ and $\mathcal{R}_{arm_2}$ that are in/near collision are invalidated (red). (c) The remaining (yellow) nodes and edges form the implicit graph spanning $\mathcal{C}_{torso} \times \mathcal{C}_{arm_1} \times \mathcal{C}_{arm_2}$. The nodes of this graph are guaranteed to be collision-free, and we can rapidly find collision-free paths following Algorithm 
    }
    \label{fig:dprm_without_and_with_collisions}
\end{figure*}

Given two DRMs ($\mathcal{R}_{arm_1}, \mathcal{R}_{arm_2}$), constructed as outlined in Algorithm \ref{alg:roadmap_construction}, how do we efficiently plan a \textit{single} collision-free path from $x_{start}$ to $x_{target} \in \mathcal{C}$? The naive approach of independently searching for a path in each DRM (e.g. using $A^*$) and then composing the two solutions would fail, because the independent paths would most likely not obey the constraint imposed by the shared torso joints. Our search algorithm is inspired by $A^*$ but must bounce between the two DRMs during its search, adding node pairs to the priority queue based on (1) satisfaction of constraints and (2) their cost in \textit{both} DRMs. The online planning algorithm is described in Algorithm \ref{alg:online_planning}, with the main workhorse being the graph search (Algorithm \ref{alg:graph_search}) within the composed roadmaps $\mathcal{R}_{arm_1}, \mathcal{R}_{arm_2}$. 

\begin{breakablealgorithm}
	\caption{Dual-Roadmap Graph Search}\label{alg:graph_search}
	\begin{algorithmic}[1]
        \STATE \textbf{Input:} $x_{start} = (x^l_s, x^r_s), ~~ x_{target} = (x^l_t, x^r_t)$
        \STATE Initialize EXPLORE list with node pair $(x^l_s,x^r_s)$ with priority 0
        \WHILE {EXPLORE list is not empty}
    	    \STATE Pop node pair $x_{c} = (x^l_c, x^r_c)$ with lowest priority from EXPLORE.
            \IF {$x_c == x_{target}$}
                \RETURN Path through graph from $x_c$ to $x_s$.
            \ENDIF
            \STATE
            \STATE \textcolor{olive}{// For all left/right neighbors of $x_c$ in $\mathcal{R}_l, \mathcal{R}_r$, compute their cost-to-go, $g$, and priority, $f$.}
            \STATE Get left/right neighbors $N_{l} = \mathcal{R}_l(x^l_c) , N_{r} = \mathcal{R}_r(x^r_c)$ 
            \FOR{$arm \in \{ l, r \}$}
    		\FOR{neighbor $x^{arm}_n \in N_{arm}$}
                \STATE Compute cost $x_n^{arm}.g$ and priority $x_n^{arm}.f$ for node $x_n^{arm}$
                \STATE ~~~~ $x_n^{arm}.g = x_c^{arm}.g + dist(x_c^{arm}, x_n^{arm})$
                \STATE ~~~~ $x_n^{arm}.f = x_n^{arm}.g + dist(x_n^{arm}, x_t^{arm})$.
            \ENDFOR
            \STATE Sort neighbors $N_{arm}$ by priority $f$. 
            \ENDFOR
            \STATE

            \STATE \textcolor{olive}{// For all (left/right) neighbors $x^a_n$, get the lowest priority (right/left) neighbor, $x^b_n$ (that satisifies torso constraints).}
            \FOR{arm pair $(a, b) \in \{(l, r), (r, l)\}$}
            \FOR{neighbor $x^a_n \in N_l$}
               \STATE Get the lowest priority $x_n^b \in N_r$ such that $Proj_{\mathcal{C}_{torso}}(x^{b}_{n}) = Proj_{\mathcal{C}_{torso}}(x^{a}_{n})$
               \STATE Define this resulting pair $x_{p} = (x^a_n, x^b_n)$.
               \STATE \textcolor{olive}{// Add resulting pair to OPEN list if (a) it hasn't been seen, or (b) we've found shorter path to it.}
               \IF {$x_p$ not in EXPLORE or $x_c.g + dist(x_c, x_p) < x_p.g$}
               \STATE Add $x_p$ to the EXPLORE list with priority $max(x_n^a.f, x_n^b.f)$.
               \STATE Store $x_c$ as the parent of $x_p$
               \ENDIF
            \ENDFOR
            \ENDFOR
		\ENDWHILE
	\end{algorithmic}
\end{breakablealgorithm}

This strategy enables sampling-based motion planning for the full 19D configuration space, while overcoming the curse of dimensionality by leveraging two roadmaps in 12D configuration spaces. The constraints imposed by the shared torso joints also enable greater efficiency in searching through the two graphs by vastly reducing the valid node pairs. The results in Section \ref{sec:results} demonstrate that we attain very fast and reliable motion planning.

As the final step (Line 11 in Algorithm \ref{alg:online_planning}), we compose paths for two arms to obtain a single path for all DoFs, and do lazy collision checking to verify the path is collision-free. If a collision is detected, the offending nodes/edges are pruned out and the graph search is repeated.

\begin{remark}
Even though the nodes of the DRM are guaranteed to be collision-free due to the node-pruning step, the \textit{edges} are not guaranteed to be collision-free. While we could theoretically include these edges in the collision map of the DRM, this would lead to a substantial increase in the size of the DRM (e.g. 100x larger). We found that this tradeoff was not worth it, as ensuring the nodes are collision-free with some padding works very well in practice. 
\end{remark}


\begin{algorithm}[tb]
	\caption{Motion Planning}\label{alg:online_planning}
	\begin{algorithmic}[1]
	    \STATE \textbf{Input:} $x_{start}, x_{target} \in \mathcal{C}$, $V \in \mathbb{V}$
	    \STATE Based on voxel map $V$, prune nodes from $\mathcal{R}_{arm_1}$ and $\mathcal{R}_{arm_2}$, such that all remaining nodes are collision-free.
		\STATE Find collision-free nodes in each DRM $x^{1}_{s} \in \mathcal{R}_{arm_1},  x^{2}_{s} \in \mathcal{R}_{arm_2}$ that are closest to $x_{start}$.
        \STATE ~ $\bullet$ ~~~~ Ensure $Proj_{\mathcal{C}_{torso}}(x^{1}_{s}) = Proj_{\mathcal{C}_{torso}}(x^{2}_{s})$
		\STATE Find collision-free nodes in each DRM $x^{1}_{t} \in \mathcal{R}_{arm_1},  x^{2}_{t} \in \mathcal{R}_{arm_2}$ that are closest to $x_{target}$.
        \STATE ~ $\bullet$ ~~ Ensure $Proj_{\mathcal{C}_{torso}}(x^{1}_{t}) = Proj_{\mathcal{C}_{torso}}(x^{2}_{t})$
		\WHILE {TRUE}
		\STATE Query roadmaps $\mathcal{R}_{arm_1}, \mathcal{R}_{arm_2}$ for free path from $x_s = (x^1_s, x^2_x)$ to $x_t = (x^1_t, x^2_t)$. \\
		\STATE ~~~~ Use Algorithm \ref{alg:graph_search} to get shortest path. \\
		\STATE Connect from $x_{start}$ to $x_s$, and from $x_t$ to $x_{target}$.
		\STATE Compose full path $\tau$ and check for collisions.
		\IF{$\tau$ is collision-free}
		\RETURN Motion plan $\tau$
		\ELSE
		\STATE Prune out nodes involving collision from $\mathcal{R}_{arm_1}$, $\mathcal{R}_{arm_2}$.
		\ENDIF
		\ENDWHILE
	\end{algorithmic}
\end{algorithm}

\section{Results}
\label{sec:results}

\begin{figure*}[htbp]
\centerline{\includegraphics[width=0.98\textwidth]{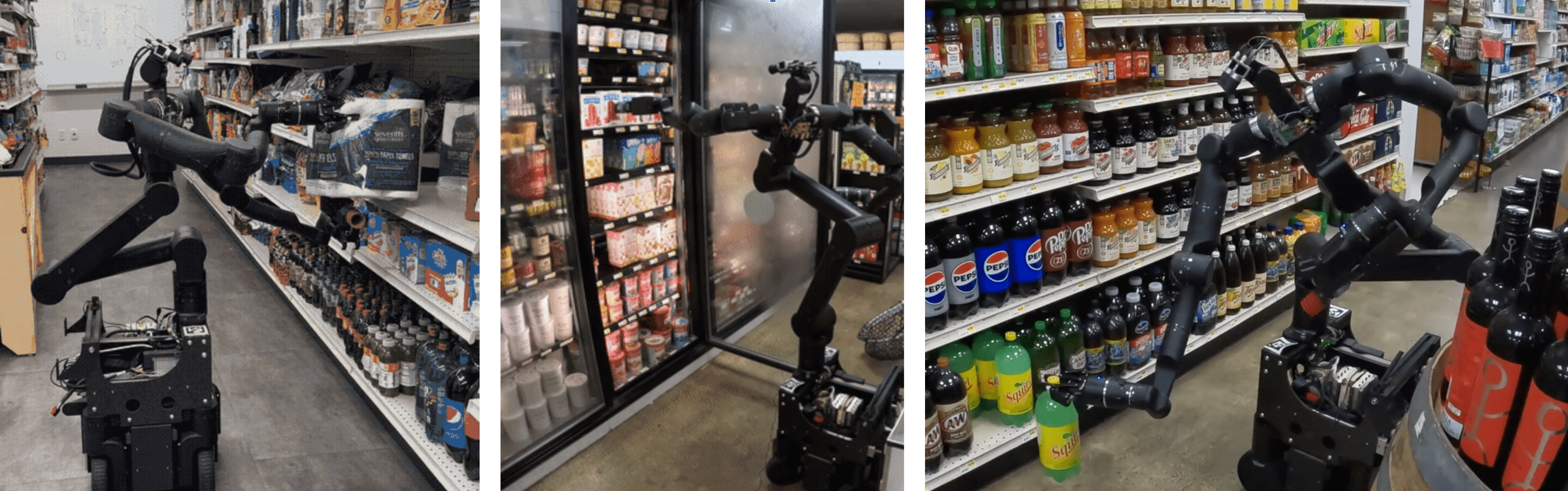}}
\caption{Snapshot images of robot motion plans while grabbing items in the grocery store.}
\label{fig:robot_motion_plans}
\end{figure*}

We demonstrate that our dual-arm motion planner works reliably and efficiently in a real-world setting through extensive robot experiments on our custom mobile manipulator. The mobile manipulator robot comprises four distinct parts:  (1) a pseudo-holonomic 4-wheeled chassis; (2) a 5-DoF torso; (3) two 7-DoF arms; and (4) a camera pair on a pan-tilt neck. For visual perception, we employ two pairs of Basler acA2500-60uc cameras mounted on the pan-tilt neck and on the chassis front. All computations are managed by a central system featuring an Intel Core i9-12900K CPU and an NVIDIA A6000 GPU. More details on the hardware setup can be found in \cite{Bajracharya2023}.

In our experiments, we tested the motion planner on the 19 DoF upper-body of the mobile manipulation robot by executing grocery fulfillment in a real grocery store located in Mountain View, CA. The robot must autonomously grab a set of unique \textit{randomly selected} items from the grocery store (out of $\approx$1000 items) and place them in a basket attached at its rear (see Figure \ref{fig:robot_image}. Therefore, the motion planner must plan motions in random parts of the grocery store as it navigates to the different items for grasping, where the items could be anywhere on a shelf/rack/hanger/etc. In some instances, the robot must open/hold a door with one arm, while planning the other arm to grab an item behind the door. Since this paper is solely focused on motion planning, we have left out further details about perception and grasping, which can be found in \cite{Bajracharya2023}.

The robot operates outside of opening hours so there was no human interaction during planning. We do not make any modifications to the grocery store; we operate in the grocery store in the condition it is left in by the store owner and customers. 

Our experiments resulted in a total of 2101 motion plans to varying targets at different parts of the grocery store (i.e. different collision environments) computed online during task execution. Our planner achieved an average planning time of 0.42s and a success rate of 99.9\% even with very challenging targets. Table \ref{table:results} summarizes these results and Figure \ref{fig:robot_motion_plans} shows snapshots during execution of a few motion plans in the grocery store. A video showing the robot running this motion planner while operating in the grocery store can be found here: \url{https://drive.google.com/file/d/1CGAwj-Cti35Q1_XXdAs26Ba-LIGW_EF4/view?usp=drive_link}.

\captionsetup{labelformat=empty}
\def\arraystretch{1.5}
\begin{table}[htbp]
\normalsize
\caption{Table 1: Experimental Results in Real Grocery Store}
\vspace{-2mm}
\begin{center}
\resizebox{0.48\textwidth}{!}{%
\begin{tabular}{|P{40mm}|P{20mm}|P{20mm}|}
\hline
Planning Time (s) & Planning Range (s) & Success Rate  \\
\hline
0.42 ~~ (avg) ~~~~~~~~~~~~~~ [0.08, 0.62] ~~ (10-90 perc.) &  [0.03, 6.19] &  ~~ 99.9\% \newline (2098 / 2101)  \\
\hline
\end{tabular}}
\label{table:results}
\end{center}
\vspace{-1mm}
\footnotesize{$^*$ Column 1 shows mean value, and also the 10-90 percentile range. Planning times are only considered for successful plans.} \\
\vspace{-3mm}
\end{table}

~

\noindent
\textbf{DRM construction and hyperparameters:} For our experiments, we built/utilized $\mathcal{R}_{arm_1}$ with 759,646 nodes and $\mathcal{R}_{arm_2}$ with 760,433 nodes. Each joint was uniformly discretized with $\pi / 6$ intervals and range varying from $[\pi / 6, \pi]$. When building the collision map, $\mathcal{M}_c$, a voxel discretization of $0.06$m was used with workspace size $2.1m \times 2.1m \times 1.9m$. Nodes were additionally pruned based on center-of-mass and tip orientation constraints. The computation time to build each of these roadmaps offline is $\approx$6hrs, and only needs to be done once per robot-embodiment.

~

\noindent
\textbf{Comparison experiments:} In addition to our extensive field tests, we ran a set of comparison experiments comparing our algorithm to the single-arm DRM approach described in \cite{Cheng2023}. The fundamental difference in our approach is that rather than adopting a leader-follower approach (with one kinematic chain utilizing the DRM, and the other arm following with a local QP-planner), we plan for both arms using the composition of two roadmaps as described in this paper. Since both planners are based on the same DRM concept, this provides a good direct comparison showcasing the value of the roadmap composition for overlapping kinematic chains.

In this set of comparison experiments, we tested our motion planner in a mock grocery store built in our lab, as it was more easily accessible. For fair comparison, the same set of unique items were requested and the same hardware was used when testing both planners. Two important differences between our real-world field test experiments (Table \ref{table:results}) and these comparison experiments (Table \ref{table:comparison_results}) is that (1) we used an Intel i9 processor for our real-world experiments versus an Intel i5 processor for our comparison experiments, and (2) the subset of picks considered in our mock grocery store were qualitatively more challenging than in the field test.

\captionsetup{labelformat=empty}
\def\arraystretch{1.5}
\begin{table}[htbp]
\normalsize
\caption{Table 2: Comparison Experiments for Motion Planner}
\vspace{-2mm}
\begin{center}
\resizebox{0.48\textwidth}{!}{%
\begin{tabular}{|P{20mm}||P{36mm}|P{19mm}|P{17mm}|}
\cline{1-4} 
~ & Planning Time (s) & Planning Range (s) & Success Rate \\
\hline
Single DRM (lead-follow) & 1.00 ~ (avg) ~~~~~~~~~ [0.18, 2.40] (10-90 perc) & [0.12, 7.53] &  ~~~93\% \newline (145 / 157)  \\
\hline
Dual Arm DRMs (ours) & 0.85 ~ (avg) ~~~~~~~~~ [0.31, 1.34] (10-90 perc) & [0.23, 9.52] & ~~~99\% \newline (155 / 157)  \\
\hline
\end{tabular}}
\label{table:comparison_results}
\end{center}
\vspace{-1mm}
\footnotesize{$^*$ Column 1 shows mean value and also the 10-90 percentile range . Planning times are only considered for successful plans.} \\
\vspace{-3mm}
\end{table}

Table \ref{table:comparison_results} summarizes the comparison experiment results. Our method outperformed the baseline method, exhibiting marginally faster average planning time and a significantly higher success rate. Our takeaways from these comparison experiments were that:
\begin{itemize}[leftmargin=*]
    \item In the nominal case, there is marginal difference in planning time between the two methods. While our method exhibits slightly faster average planning times, this is well within the variance of planning times. In fact, the median planning times were 0.63s for the single DRM and 0.61s for the Dual Arm DRMs (only a 0.02s difference).
    \item The value of our dual-DRM approach is primarily shown in more ``challenging'' motion plans, particularly where the two arms might come close. This is reflected in the higher success rate. This also led to many more ``long'' motion plans for the single DRM approach, which was the primary cause of the longer average planning times for that method.
\end{itemize}


\section{Conclusion}

By leveraging parallel and/or offline computation, sampling-based motion planners have become very successful in a wide array of robotic applications due to their simplicity, robustness, and ability to deal with changing environments. However, they inevitably suffer from the curse of dimensionality as we increase the dimensionality of the configuration space. This paper leverages the structure introduced by dual arm manipulators to help alleviate the curse of dimensionality by intelligent composition of per-(kinematic) chain dynamic roadmaps. We derive a novel motion planning algorithm based on this concept, and demonstrate its effectiveness on a dual-arm mobile manipulation platform.



\bibliographystyle{IEEEtran}
\bibliography{references}

\section{Appendix}
\label{sec:appendix}

\setcounter{proposition}{0}
\begin{proposition}
    A node pair $(x_{\mathcal{R}_1}, x_{\mathcal{R}_2})$ (where $x_{\mathcal{R}_i} \subset \textnormal{supp}(\mathcal{R}_{arm_i})$) is in collision if and only if at least one of the following holds:
    \begin{itemize}
        \item $(x_{\mathcal{R}_1}, x_{\mathcal{R}_2})$ is contained in the inter-arm collision maps $\mathcal{M}_{inter_1}$ or $\mathcal{M}_{inter_2}$,
        \item $x_{\mathcal{R}_1}$ is in collision,
        \item $x_{\mathcal{R}_2}$ is in collision.
    \end{itemize}
\end{proposition}

\begin{proof}
For the node pair $x = (x_{\mathcal{R}_1}, x_{\mathcal{R}_2})$, physically $x_{\mathcal{R}_1}$ represent the configurations of kinematic chain 1 and $x_{\mathcal{R}_2}$ represents the configurations of kinematic chain 2. Let us define robot collision by enumerating all the conditions under which these kinematic chains are in collision; a collision results if and only if at least one of the following holds:
\begin{enumerate}
    \item a collision between kinematic chains (inter-arm),
    \item a self-collision within kinematic chain 1,
    \item a self-collision within kinematic chain 2,
    \item a collision between kinematic chain 1 and a voxel,
    \item a collision between kinematic chain 2 and a voxel.
\end{enumerate}

Let us first prove the first direction (that if any of the 3 conditions holds, then there is a collision).
\begin{itemize}
    \item Case $(x_{\mathcal{R}_1}, x_{\mathcal{R}_2}) \in \mathcal{M}_{inter_1} \cup \mathcal{M}_{inter_2}$ : By definition, if the node pair is contained in the inter-arm collision map, then there is a collision between the arms, fulfilling condition 1 for collision above.
    \item Case $x_{\mathcal{R}_1}$is in collision: By definition, either condition 2 or 4 above is fulfilled implying robot collision.
    \item Case $x_{\mathcal{R}_2}$ is in collision: By definition, either condition 3 or 5 above is fulfilled implying robot collision.
\end{itemize}
Therefore, it is trivially shown that if any of the 3 conditions hold, there is a collision.

Now let us prove the other direction (that if the node pair is in collision, the at least one of the 3 conditions must hold). Assume, for contradiction, that the node pair $x = (x_{\mathcal{R}_1}, x_{\mathcal{R}_2})$ is in collision, but none of the three conditions hold.

Stepping through each of these conditions, we outline the implications of each of them \textit{not} holding.
\begin{itemize}
    \item If $(x_{\mathcal{R}_1}, x_{\mathcal{R}_2}) \notin \mathcal{M}_{inter_1} \cup \mathcal{M}_{inter_2}$, then condition 1 for robot collision outlined above is \textit{not} satisfied.
    \item If $x_{\mathcal{R}_1}$ is not in collision, then conditions 2 and 4 for robot collision are not satisfied.
    \item If $x_{\mathcal{R}_2}$ is not in collision, then conditions 3 and 5 for robot collision are not satisfied.
\end{itemize}
Therefore, if none of the three conditions in the proposition hold, then none of the 5 conditions required for robot collision are satisfied, and the node pair $x = (x_{\mathcal{R}_1}, x_{\mathcal{R}_2})$ cannot be in collision. This is a contradiction. This proves that if the node pair $x = (x_{\mathcal{R}_1}, x_{\mathcal{R}_2})$ is in collision, then at least one of the 3 conditions in the proposition must hold.
\end{proof}

\end{document}